\documentclass[runningheads]{llncs}

\usepackage{amsmath,amsfonts,amssymb}
\usepackage{graphicx,xcolor}
\usepackage{booktabs}
\usepackage{authblk}
\usepackage{todonotes}
\usepackage{bm}
\usepackage[numbers]{natbib}
\usepackage{hyperref}

\begin{document}

\title{Evolutionary Planning in Latent Space}

\author{Thor V.A.N. Olesen\thanks{Contributed equally} \and Dennis T.T. Nguyen$^{*}$ \and Rasmus B. Palm \and Sebastian Risi}


\authorrunning{Olesen, Nguyen, Palm,  Risi}

\institute{IT-University of Copenhagen}

\maketitle

\begin{abstract}
Planning is a powerful approach to reinforcement learning with several desirable properties. However, it requires a model of the world, which is not readily available in many real-life problems.
In this paper, we propose to learn a world model that enables \emph{Evolutionary Planning in Latent Space} (EPLS). We use a Variational Auto Encoder (VAE) to learn a compressed latent representation of individual observations and extend a Mixture Density Recurrent Neural Network (MDRNN) to learn a stochastic, multi-modal forward model of the world that can be used for planning. We use the Random Mutation Hill Climbing (RMHC) to find a sequence of actions that maximize expected reward in this learned model of the world.
We demonstrate how to build a model of the world by bootstrapping it with rollouts from a random policy and iteratively refining it with rollouts from an increasingly accurate planning policy using the learned world model. 
After a few iterations of this refinement, our planning agents are better than standard model-free reinforcement learning approaches demonstrating the viability of our approach\footnote{Code to reproduce the experiments are available at \url{https://github.com/two2tee/WorldModelPlanning}} \footnote{Video of driving performance is available at \url{https://youtu.be/3M39QgeF27U}}.
\end{abstract}


\begin{figure}[h!]
    \centering
    \includegraphics[width=1.0\linewidth]{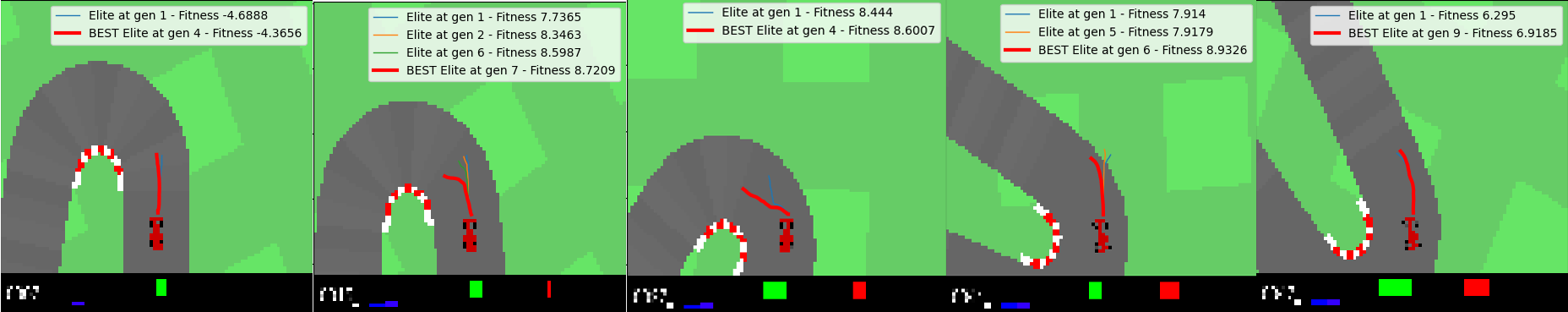}
    \caption{Planed trajectories using evolutionary planning in the latent space of a learned world model.}
    \label{fig:rmhc_s_turn_planning_trajectory}
\end{figure}

\section{Introduction}
Planning---searching for action sequences that maximize expected reward---is a powerful approach to reinforcement learning problems which has recently lead to breakthroughs in hard domains such Go, Shogi, Chess, and Atari games \cite{silver2016mastering, silver2017mastering, schrittwieser2019mastering}. To plan, the agent needs access to a model of the world which it can use to simulate the outcome of actions, to determine which course of action is best. Planning using a model of the world also allows you to introspect what the agent is planning and why it thinks a certain action is good. It even allows you to add constraints or change the objective at runtime.

For some domains, these world models are readily available, for instance, games, where the world model is given by the rules of the game. However, for many real-world problems, e.g. driving a car, they are not available.

In problems where a model of the world is not available one can instead use model-free reinforcement learning, which learns a policy that directly maps from the environment state to the actions that maximize expected reward. However, these approaches require a large number of samples of the real environment---which is often expensive to obtain---to learn the statistical relationship between states, actions, and rewards. This is especially true if the environment requires complex sequences of actions before any reward is observed.

Alternatively, a model of the world can be \textit{learned}, which is the approach taken in this paper. This is known as model-based reinforcement learning. Learning a model of the world often requires much fewer samples of the environment than directly learning a policy since it can use supervised learning methods to predict the environment state transitions, whether or not the reward signal is sparse. Several models have been proposed for learning world models \cite{worldmodels, schrittwieser2019mastering, hafner2019learning}.

In this paper, we propose an extension to the Mixture Density Recurrent Neural Network (MDRNN) world model \cite{worldmodels}, which makes it suitable for planning, and demonstrate how we can do evolutionary planning entirely in the latent space of the learned world model. See figure \ref{fig:rmhc_s_turn_planning_trajectory} for examples of planned trajectories and figure \ref{fig:plan_loop} for an overview of the proposed method.

We further show how to iteratively improve the world model by using the existing world model to do the planning and using the planning policy to sample better rollouts of the environment, which in turn are used to train a better world model. This process is bootstrapped by using an initial random policy. We show that after only a few rounds of iterative refinement like this we achieve results that are better than standard model-free approaches, demonstrating the viability of the approach.

\section{Related work}

\subsection{Planning}
In planning an agent uses a model of the world to predict the consequences of its actions and select an optimal action sequence accordingly. Planning is a powerful technique that has recently lead to breakthroughs in hard domains such as Go, Chess, Shogu, and Atari \cite{silver2016mastering, silver2017mastering, schrittwieser2019mastering}. 

Monte-Carlo Tree Search (MCTS) is a state-of-the-art planning algorithm for discrete action spaces, which iteratively builds a search tree that explores the most promising paths using a fast, often stochastic, rollout policy \cite{browne2012survey}. 

Rolling Horizon Evolutionary Algorithms (RHEA) encode individuals as sequences of actions and uses evolutionary algorithms to search for optimal trajectories. \textit{Rolling Horizon} (RH) refers to how the first action of a plan is executed before the plan is reevaluated and adjusted, looking one step further into the future and slowly expanding the horizon \cite{perez2013rolling, gaina2017population}. RHEA naturally handles continuous action spaces. The authors in \cite{tong2019enhancing} show how to learn a prior for RHEA by training a value and policy network. The value network reduces the required planning horizon by estimating the rewards of future states. The policy network helps initialize the population of planning action trajectories to help narrow down the search scope to a near-optimal local action policy-subspace. In our approach, we use a randomly initialized set of planning trajectories that are improved iteratively with evolution. 

Random Mutation Hill-Climb (RMHC) is a simple and effective type of evolutionary algorithm that repeats the process of randomly selecting a neighbour of a best-so-far solution and accepts the neighbour if it is better than or equal to the current best-so-far solution. This local search method starts with a solution and iteratively tries to improve it by taking random steps or restarting from another region in the policy space. 

Planning approaches that rely on imperfect models may plan non-optimal trajectories. The authors of \cite{ovalle2020bootstrapped} suggest incorporating uncertainty estimation into the forward model, which enables the agent to perform better. In general planning under uncertainty has been extensively studied \cite{blythe1999overview, michie1966game, kahn2017uncertainty}.

\subsection{Learning world models}

If a world model is not available it can be learned from observations of the environment. This is generally known as model-based Reinforcement Learning (RL).

\textit{World Models} \cite{worldmodels} introduces a stochastic recurrent world model which is learned from observations of the environment under an initially random policy. The model uses a Variational Auto-Encoder (VAE) to encode pixel inputs into a low dimensional latent vector. A recurrent neural network (RNN) is then trained to predict sequences of these latent vectors using a Gaussian Mixture Model to capture the uncertain and multi-modal nature of the environment. Notably, the authors do not use this world model for planning, but rather for training a relatively simple policy network.

In \textit{MuZero} \cite{schrittwieser2019mastering}, the authors show how to do planning with a learned model in board and video games using tree-based search (MCTS) to enable imitation learning with a learned policy network. 

In \textit{PlaNet} \cite{hafner2019learning}, the authors have shown it is possible to do online planning in latent space using an adaptive randomized algorithm on a recurrent state-space model (SSM) with a deterministic and stochastic component and a multi-step prediction objective. \textit{PlaNet} \cite{hafner2019learning} is the approach that is most similar to the work presented here (i.e., online planning on a learned model). However, it uses a rather complicated dynamics model and planning algorithm. In \textit{Dreamer} \cite{hafner2019dream}, the authors use the \textit{PlaNet} world model but no longer do online planning. Instead, their Dreamer agent uses an actor-critic approach to learn behaviors that consider rewards beyond a horizon. Namely, they learn an action model and value model in the latent space of the world model. Thus, their approach is similar to \textit{World Models} \cite{worldmodels} where they plan on a learned model by training a policy inside the simulated environment with backpropagation and gradient descent, instead of evolution. The novel part is using a value network to estimate rewards beyond a finite imagination horizon. Also, \textit{World Models} \cite{worldmodels} does not show how to do planning on a fully learned model, since the reward signal is not learned in their model. Finally, MuZero \cite{schrittwieser2019mastering} relies on extensive training data and access to unrealistic GPU resources, which may not be feasible in practice. 

In another related approach, \textit{Neural Game Engine} \cite{neuralgameengine}, the authors show how to learn accurate forward models from pixels that can generalize to different size game levels. However, their methods currently only work on grid-based world games. The authors argue it does not work as a drop-in replacement for the kind of world models we need in real-life environments. For this purpose, the authors recommend looking into some of the previously presented methods that learn a latent dynamics model with a 2D state space model (SSM) like shown in PlaNet and Dreamer that both use a Recurrent State Space Model (RSSM).   

\section{Approach} \label{approach_section}
\begin{figure}[!ht]
\centering
\includegraphics[width=\linewidth]{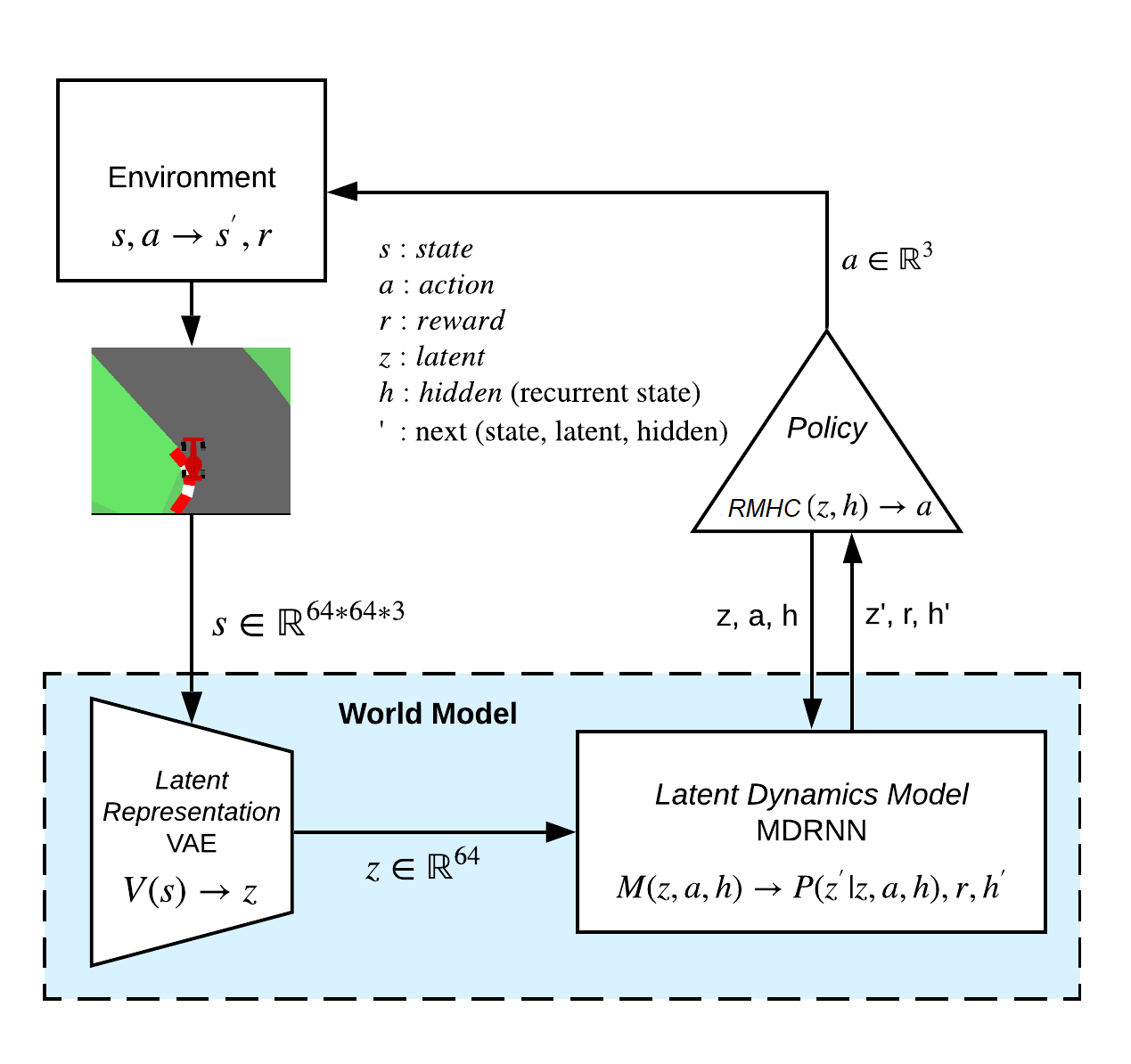}
\caption{\textbf{Evolutionary Planning in Latent Space (EPLS)}. The raw observation is compressed by V at each time step $t$ to produce a latent vector $z_t$. RMHC does planning by repeatedly generating, mutating, and evaluating action sequences $a_0,...,a_T$ in the learned world model, M. where $T$ is the horizon. The learned world model, M, receives an action $a_t$, latent vector $z_t$ and hidden state $h_t$ and predicts the simulated reward $r_t$, next latent vector $z_{t+1}$, and next hidden state $h_{t+1}$. The predicted states are used with the next action as inputs for M to let the agent simulate the trajectory in latent space. The first action of the plan with the highest expected total reward in the simulated environment is executed in the real environment.}
\label{fig:plan_loop}
\end{figure}

We use a model-based RL approach to solve a continuous reinforcement learning control task. We achieve this through online evolutionary planning on a learned model of the environment. Our solution combines a world model \cite{worldmodels} with rolling horizon evolutionary planning \cite{perez2013rolling}.  See figure \ref{fig:plan_loop} for an overview.

Similar to the model in the original world model \citep{worldmodels}, our model uses a visual sensory component (V) to compress the current state into a small latent representation. We extend the memory component (M) so that it predicts the next latent state, the expected reward, and whether the environment terminates. The decision-making component in the original world model \citep{worldmodels} uses a simple learned linear model that maps latent and hidden states directly to actions at each time step. In contrast, our work uses a random mutation hill-climbing (RMHC) planning algorithm as the decision-making component that exploits M to do online planning in latent space.

\subsection{Learning the World Model}

\begin{figure}[ht]
    \centering
    \includegraphics[width=\linewidth]{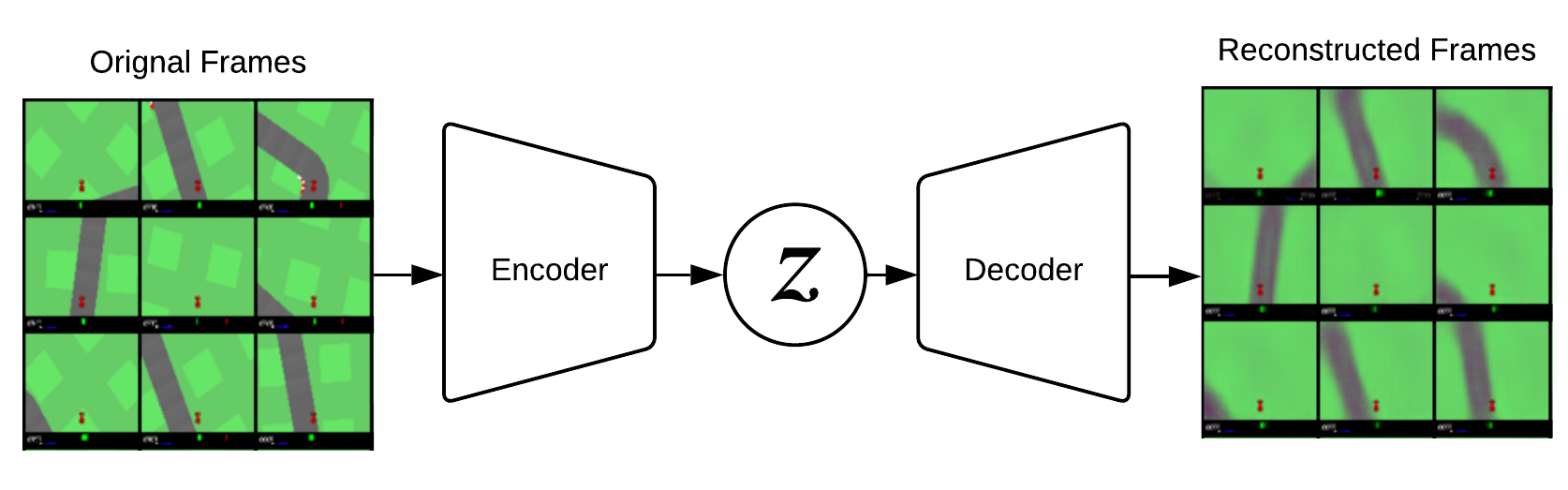}
    \caption{Flow diagram of a Variational Autoencoder (VAE). The VAE learns to encode frames into latent vectors by minimizing the pixel-wise difference between input frames and reconstructed frames (i.e., L2 or MSE) generated by decoding the latent vectors.}
    \label{fig:vae_encoding_decoding}
\end{figure}

The \textbf{visual component} (V) is implemented as a convolutional variational autoencoder (ConvVAE), which learns an abstract, compressed representation $z_t \in \mathbb{R}^{64}$ of states (i.e., frames) $s_t \in \mathbb{R}^{64 \times 64 \times 3}$ using an encoder and decoder as shown in figure~\ref{fig:vae_encoding_decoding}. 

The VAE encoder is a neural network that outputs a compressed representation of a state $s$ (frame) using a deep convolutional neural network (DCNN) of four stacked convolutional layers and non-linear relu activations to compress the frame and two fully-connected (i.e., dense) layers that encode the convolutional output into low dimension vectors $\mu_z$ and $\sigma_z$: 
\begin{equation}
    encoder: s \in \mathbb{R}^{64 \times 64 \times 3} \rightarrow \mu_z \in \mathbb{R}^{64}, \sigma_z \in \mathbb{R}^{64} \,.
\end{equation}
The means $\mu_z$ and standard deviations $\sigma_z$ are used to sample a latent state $z$ from a multivariate Gaussian with diagonal covariance:
\begin{equation}
    z \in \mathbb{R}^{64} \sim \mathcal{N}(z | \mu_z, \sigma_z) \,.
\end{equation}

The decoder is a neural network that learns to decode and reconstruct the state (i.e., frame) $s$ given the latent state $z$ using a deep CNN of four stacked deconvolution layers: 
\begin{equation}
    decoder: z \in \mathbb{R}^{64} \rightarrow s' \in \mathbb{R}^{64 \times 64 \times 3} \,.
\end{equation}
Each convolution and deconvolution layer uses a stride of two. Convolutional and deconvolutional layers use relu activations. The output layer maps directly to pixel values between 0 and 1. The VAE is trained with the standard VAE loss \cite{kingma2013auto}.

We extend the \textbf{memory component} (M) of \cite{worldmodels} to also output an expected reward $r$ and a binary terminal signal $\tau$ to obtain a fully learned world model that can be used for planning entirely in latent space. M is an LSTM with 512 hidden units, which jointly models the next latent state $z_{t}$, reward $r_{t}$ and whether or not the environment terminates, $\tau_{t}$,

\begin{equation}
    p(z_{t}, r_{t}, \tau_{t} | h_{t-1}) = p(z_{t} | h_{t-1}) p(r_{t} | h_{t-1}) p(\tau_{t} | h_{t-1}) \,.
\end{equation}

The LSTM hidden state $h_t$ depends on the previous hidden state $h_{t-1}$, the current action $a_t$, and the current latent state $z_t$ such that $h_t = \text{LSTM}(z_t, a_t, h_{t-1})$.

Most complex environments are stochastic and multi-modal so $p(z_{t}|h_{t-1})$ is approximated as a mixture of Gaussian distribution (MD-RNN). The output of the MDRNN are the parameters $\pi, \mu, \sigma$ of a parametric Gaussian mixture model where $\pi$ represents mixture probabilities:

\begin{equation}
    p(z_{t} | h_{t-1}) = \sum_{k=1}^5 \pi_k \mathcal{N}(z_{t} | \mu_k, \sigma_k) \,,
\end{equation}
where $\pi$, $\mu$ and $\Sigma$ are linear functions of $h_{t-1}$ and each mixture component is a multivariate Gaussian distribution with diagonal covariance.

We model the reward $r$ using a Gaussian with a fixed variance of 1 such that
\begin{equation}
    p(r_{t}| h_{t-1}) = \mathcal{N}(r_{t} | \mu^\tau_{t}, 1) \,, 
\end{equation}
where $\mu^\tau_{t}$ is a linear function of $h_{t-1}$. Finally we model the terminal state $\tau$ using a Bernoulli distribution,
\begin{equation}
    p(\tau_{t} | h_{t-1}) = p^{\tau_{t}}(1-p)^{1-\tau_{t}} \,,
\end{equation}
where $p = \text{sigmoid}(f(h_{t-1}))$ is the sigmoid of a linear function of $h_{t-1}$.

We train M by minimizing the negative log-likelihood of $p(z_{t}, r_{t}, \tau_{t} | h_{t-1})$ for observed rollouts of the environment,
\begin{equation}
    \mathcal{L} = -\log p(z_t, r_t, \tau_t | h_{t-1}) = \text{MSE}(r_t, \hat{r}_t) + \text{BCE}(\tau_t, \hat{\tau}_t) + \text{GMM-NLL}(z_t, \hat{z_t}) \,,
\end{equation}
where MSE is the mean squared error, BCE is the binary cross-entropy and GMM-NLL is the negative log likelihood of a gaussian mixture model and $\hat{z}, \hat{r}, \hat{\tau}$ are the observed latents, reward and termination state.

\subsection{Evolutionary planning in latent space}

\begin{figure}[ht!] 
        \centering
        \includegraphics[width=\linewidth]{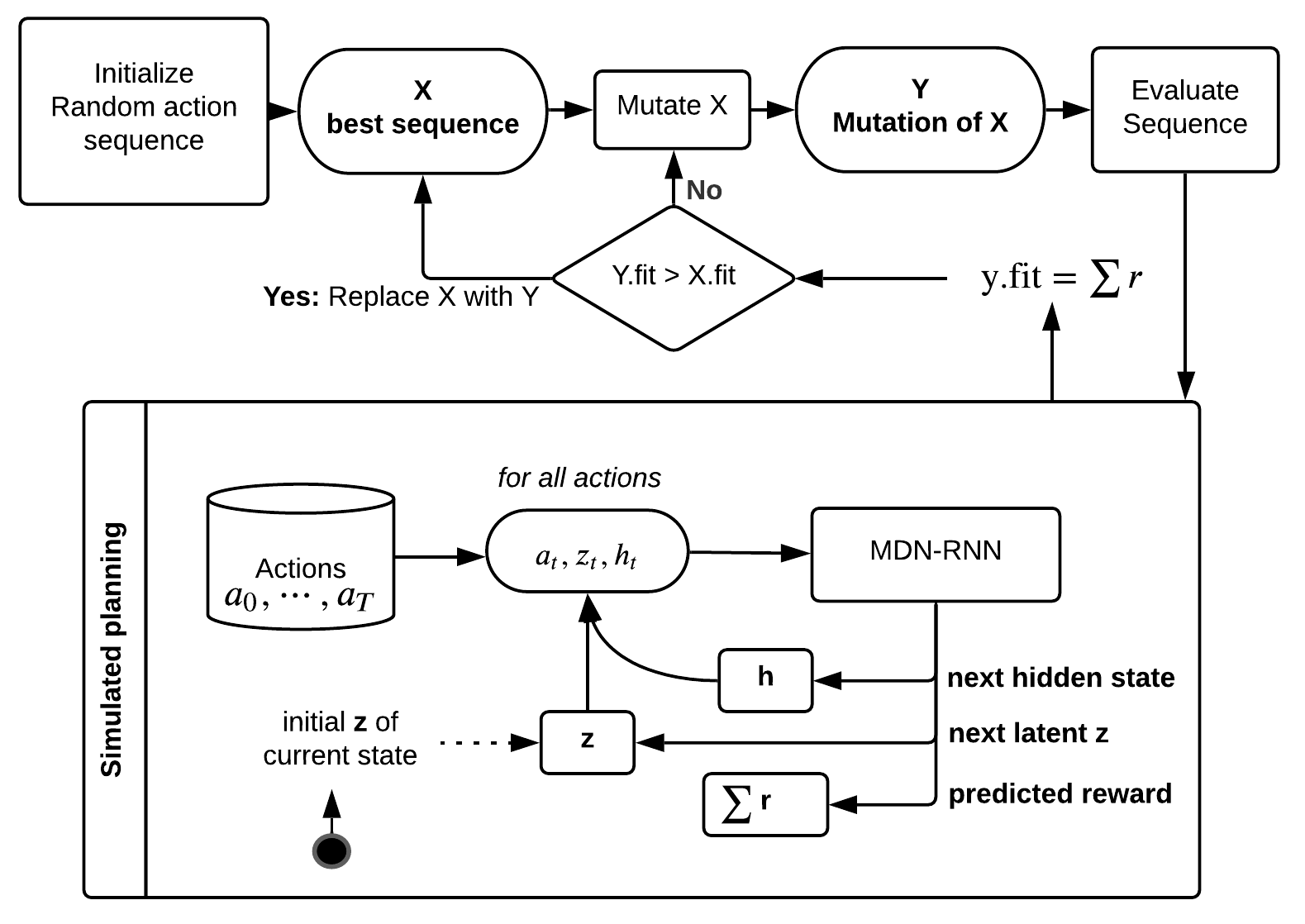}
        \caption{\textbf{Planning details.} RMHC initializes a random sequence of actions sampled from the environment and mutates it repeatedly across generations. Each plan is evaluated in latent space using the simulated environment where the fitness metric is the total undiscounted expected reward associated with executing the planning trajectory in latent space.}
        \label{fig:eval_individual}
\end{figure}

Once the world model is trained it can be used for planning. We use Random Mutation Hill Climbing (RMHC) which is a simple evolutionary algorithm. RMHC works by iteratively mutating and evaluating individuals, and letting the elite be the basis for the next round of mutation. We use RMHC to find a sequence of actions that maximize the expected reward as predicted by the world model. The length of the action sequence determines how far into the future the agent plans and is known as the horizon. Finally, we use shift buffering to avoid repeating the entire search process from scratch at every time step \cite{gaina2020rolling}. In short, after each planning step we pop the first action of the action sequence and add a new random action to the end of the action sequence. This modified plan is then the starting point for the next planning step. See figure \ref{fig:eval_individual} for an overview.

\section{Experiments}
We test our approach on the continuous control \texttt{CarRacing-v0} domain \cite{carracing}, built with the Box2D physics engine. At every trial, the agent is exposed to a randomly generated track. Reaching a high score requires the agent to plan how to make each turn with continuous actions, which makes it a suitable test domain for our evolutionary latent planning approach. The environment yields a reward of -0.1 each time step and a reward of +1000/$N$ for each visited track tile where $N$ is the total number of tiles in the track. While it is not necessarily difficult to drive slowly around a track, reaching a high reward is difficult for many current RL methods \cite{worldmodels}.

Since the environment gives observations as high dimensional pixel images, these are first resized to $64 \times 64$ pixels. The resized images are used as observations in our world model. Pixels are stored as three floating-point values between 0 and 1 that represent each of the RGB channels. The dimension of our latent space is $64$ since this yielded better reconstructions than using $32$ as in \citep{worldmodels}. Actions contain three numeric components that represent the degree of steering, acceleration, and braking.

\subsubsection*{Capturing rollouts}
The MDN-RNN and VAE models are trained in a supervised manner, which relies on access to a representative dataset of environment rollouts for training and testing. Each sample is a rollout of the environment and consists of a sequence of (state, action, reward, terminal) tuples. The state, reward, and terminal are produced by the environment given an action, which is produced by the policy. We initially use a random policy on the environment and record states, rewards, actions, and terminals for $T$ steps. We use $T=500$ for the non-iterative procedure and $T=250$ for the iterative procedure. We found that using $T=250$ was sufficient while speeding up the iterative training procedure.

\subsubsection*{Non-iterative training procedure}
The non-iterative training procedure follows the same approach as presented in the original world model work \citep{worldmodels}. To train the VAE and MDN-RNN, we first collect a dataset of 10,000 rollouts using a random policy to explore the environment where we record the random action $a_t$ executed and the generated observations. The dataset is used to train the VAE so it can learn an abstract and compressed representation of the environment. The VAE is trained for 50 epochs with a learning rate of $1e-4$ using the \textit{Adam} optimizer.

The MDN-RNN is trained on the 10,000 rollouts where each frame $s_t$ is pre-processed by the VAE into latent vector $z_t$ for each time step $t$. The latent vectors and actions $a_t$ are given to the MDN-RNN such that it can learn to model the next latent vector $p(z_{t+1} | a_t, z_t, h_t)$ as a mixture of Gaussians and the reward $r$ and the terminal $d$. The MDN-RNN consists of 512 hidden units and a mixture of 5 Gaussians. We train the MDN-RNN for 60 epochs with a learning rate of $1e-3$ using the \textit{Adam} optimizer. In summary, the full non-iterative training procedure is shown below: 

\begin{enumerate}
    \item Collect 10.000 rollouts with a random policy
    \item Train world model using random rollouts.
    \item Evaluate the agent on 100 random tracks using the RMHC planning policy.
\end{enumerate}

\subsection*{Iterative training procedure}
Once the world model is trained non-iteratively, we can use it in conjunction with our planning algorithm (RMHC) to do online planning. However, while the agent may be able to somewhat stay on the road and drive slowly at corners, its performance is limited by our world model that is trained with a random policy only. Consequently, the dynamics associated with well-behaved driving might be underexplored, and hence our world model may not be able to fully represent this in latent space.

To address this we used an iterative training procedure as suggested in \cite{worldmodels}, in which we iteratively collect rollouts using our agents planning policy and improve our world model (and thus our planning) using the new rollouts. Intuitively, we expect planning using the learned world model to yield a better policy than a random one. These new rollouts using planning are stored in a replay buffer that contains both old and new rollouts, which allows the MDN-RNN to learn from both past and new experiences. We collect 500 rollouts per iteration. The iterative training procedure is as follows:
\begin{enumerate}
    \item Train MDN-RNN and VAE non-iteratively to obtain baseline model
    \item Collect rollouts using RMHC planning policy and add them to the replay buffer
    \item Train the world model using rollouts in replay buffer.
    \item Evaluate the agent on 100 random tracks using RMHC planning policy.
    \item Go back to (2) and repeat for $I$ iterations or until the task is complete
\end{enumerate}

For both approaches, we found that training the VAE using 10k random rollouts was sufficient in representing different scenarios of the car racing environment across all our experiments. We used RMHC with a horizon of 20, and the action sequence was evolved for ten generations at every time step $t$ with shift buffering. 
\section{Results} \label{results}



\subsection{Non-iterative training}
The MDN-RNN serves as a predictive model for future latent  $z$ vectors that the VAE may produce and the expected reward $r$ that the environment may produce. Thus the rollouts used to train the MDN-RNN may affect its predictive ability and how well it represents the real environment during online planning. Thus, we trained two MDN-RNNs using the non-iterative training procedure to obtain a random model and an expert model. The random model is trained on 10,000 random rollouts and acts as our baseline model for all iteratively-trained models. The expert model trains on 5,000 random rollouts and 5,000 expert rollouts. The expert rollouts are collected with the pre-trained agent in \textit{World Models} \cite{worldmodels}. The random rollouts allow the MDN-RNN to learn the consequences of bad-driving behavior, and the expert rollouts allow it to learn the positive reward signal associated with expert-driving. The expert model is a reference model used for comparison with our agent that helps determine how well an agent may perform when the MDN-RNN is exposed to a well-representative dataset. 

\begin{figure}[ht!]
     \centering
     \includegraphics[width=0.7\linewidth]{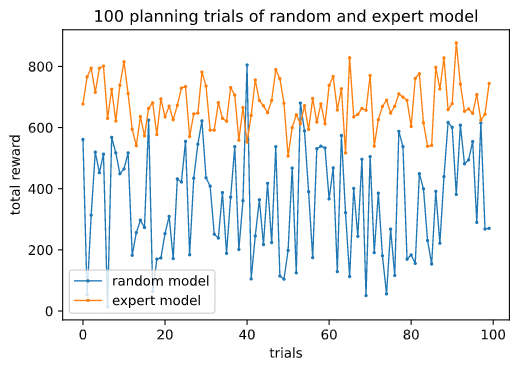}
     \caption{Total rewards of 100 trials with MDN-RNN trained on 10000 rollouts using random policy vs. an MDN-RNN trained on 5000+5000 rollouts using random and expert policy respectively. The latter yields a much higher total reward due to the dataset containing a mixture of random and well-behaved rollouts.}
     \label{fig:expert_vs_random}
\end{figure}

Using the random model, the agent did learn to drive unsteadily around the track and sometimes plan around sharp corners. However, the agent only managed to achieve a mean score of $356.20 \pm 176.69$ with the highest score of $804$. In contrast, using the expert model, the agent managed to obtain a mean score of  $765.17 \pm 102.18$ with the highest score of $900$. The expert rollouts improve the MDN-RNN's ability to capture the dynamics of the environment, which significantly improved the agent's performance (Figure~\ref{fig:expert_vs_random}).

\subsection{Iterative training}

\begin{figure}[ht!]
    \centering
    \includegraphics[width=0.7\linewidth]{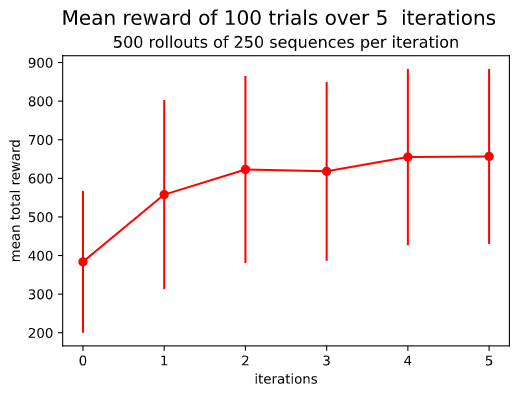}
    \caption{Mean total rewards and standard deviations over 100 trials across five iterations. The 0th iteration represents the mean total reward before iterative training, but after training on 10000 rollouts obtained from a random policy. Notice, using 500 rollouts (Experiment A) yields a faster training time and a better result compared to experiment B that uses 10.000 rollouts per iteration}
    \label{fig:experiment_iter}
\end{figure}

Since we cannot rely on access to pre-trained expert rollouts, we have implemented an iterative training procedure that allows the agent to improve its performance over time by learning from its own experiences. Namely, we generate rollouts by online planning with the RMHC evolutionary policy search method, which iteratively improve our world model. 
We investigate if the random baseline model can improve by using a small number of only 500 rollouts and a sequence length of 250 experiences trained over ten epochs and five iterations. Figure~\ref{fig:experiment_iter} shows the mean total rewards after each of the five iterations. 

Already after a single training iteration, the agent managed to get a mean score of $557.87 \pm 244.97$ and peaked at iteration 5 with a mean score of $656.82 \pm 226.67$. Despite not beating the expert model, we saw improvements throughout the iterations, and the agent managed to occasionally complete the game by scoring a total reward of 900 during benchmarks. While the first iteration yielded the most significant improvement in total average reward, the following iterations still improved. The large improvement seen in the first iteration might be due to the MDN-RNN learning the dynamics of more well-behaved driving from the agent's planning policy, which ultimately mitigates the errors made by the initially random model. 

\subsubsection*{Investigating different planning horizons and generations}

The benchmarks from iterative training show how refining the world model can affect the agent's planning capabilities. Given the best iterative model found after five training iterations, it is interesting to see how different horizon lengths and max generations affect the agent's ability to plan with the iterative model and the RMHC policy search method. 



\begin{figure}[ht!]
    \centering
    \includegraphics[width=1.0\linewidth]{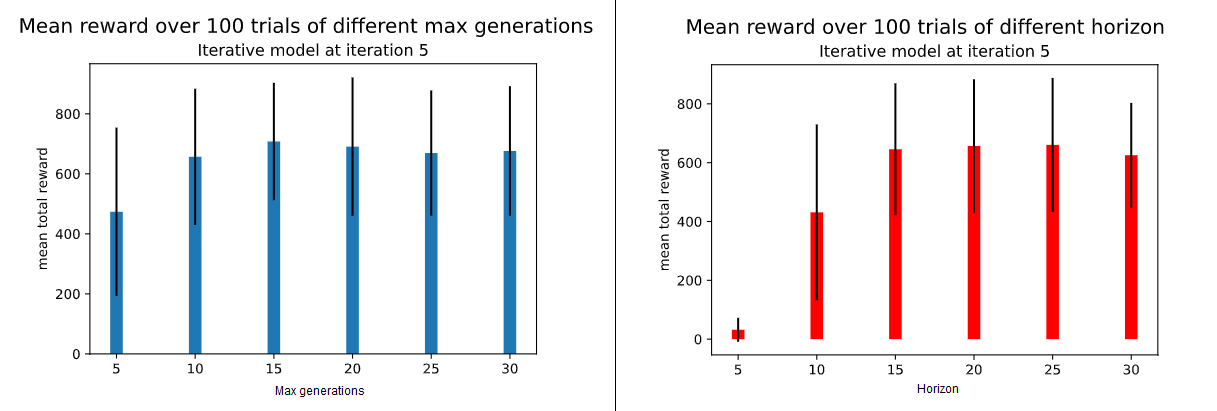}
    \caption{Left: max planning generations vs. mean rewards. Right: Horizon planning length vs. mean reward. While a  minimum number of generations and horizon length are necessary for the agent to plan  well, increasing these values further does not increase the performance of the agent.}
    \label{fig:horgendiff}
\end{figure}

Both planning parameters are adjusted independently and individually to see how they affect planning. However, we must keep in mind that the horizon length and the maximum number of generations are very likely to be highly correlated. Thus, one should also conduct experiments where both parameters are adjusted together. Figure \ref{fig:horgendiff} shows how different parameter values affect the average total reward obtained by planning with RMHC on a model trained with five iterations across 100 trials.  


The baseline number of generations is 10. Reducing this to 5 shows a decrease in mean total reward, achieving a score of  $473.53 \pm 280.18$. This reduction is likely due to the agent having less planning time and therefore not being able to converge to a better trajectory in a local policy subspace. Increasing the number of generations to 15 increases the mean reward to $707.79 \pm 195.44$, which is not surprising since it gives the agent more planning time. However, increasing the number of generations further did not improve the results. Presumably, this may imply that the agent has converged to a locally optimal trajectory in the simulated environment after being evolved 15 generations.

The results when varying the horizon are shown in Figure~\ref{fig:horgendiff}, right. The baseline horizon length is 20. Reducing this value to 5 resulted in poor planning and yielded a mean score of $31.91\pm40.54$. Seemingly, a small horizon exploits a more certain near-future but does not bring much information for long-term planning of a trajectory associated with well-behaved driving. Consequently, this kind of short-sighted agent may not be able to act in time before driving into the grass or know in what direction it should move. As we increase the horizon from 5 to 20, the mean score increases. The agent receives more information about the car's trajectory, which allows it to plan accordingly. However, a horizon beyond 20 does not help the agent, which is likely due to the increased uncertainty caused by planning too far into the future. The further the agent plans ahead, the more uncertain the trajectory becomes, which makes it less relevant to the current situation that the agent must act upon. 

According to figure \ref{fig:horgendiff} and table \ref{tabel:results}, the main results are as follows. Firstly, the iterative training procedure significantly improved our random baseline model and showed improvement after only one iteration. Secondly, increasing the maximum number of generations to 15 and a horizon of 20 used in our RMHC policy search approach improved the total average reward obtained across 100 random tracks. However, increasing the parameter values more than this yields diminishing returns, and a slight decrease in total reward. This may be due to the model's inability to predict far into the future when using a high horizon or the planning trajectory having converged when using a large number of generations. Notice, we do not dynamically adjust the horizon and number of generations during iterative training but keep them fixed during all five iterations. Instead, we compare different combinations of parameters across whole runs of five iterations. To sum up, our results show it is possible to beat traditional model-free RL methods with an evolutionary online planning approach, although we are not yet able to consistently beat or match the learned expert model presented in \textit{World Models} \cite{worldmodels}.  

\begin{table}[!ht] \label{tabel:results}
    \centering
    \begin{tabular}{l|r}
        Methods & Mean scores \\\hline
        DQN \citep{dqnresults} & $343 \pm 18$ \\
        \textbf{Non-Iterative Random Model} & $ \bm{356 \pm 177} $ \\
        A3C (Continuous) \citep{jang2017reinforcement}  & $591 \pm 45$ \\
        \textbf{Iterative Model (5 iterations, 15 gen., 20 horizon) } & $ \textbf{708} \pm \textbf{195} $ \\
        \textbf{Non-Iterative - Expert Model} & $ \textbf{765} \pm \textbf{102} $  \\
        World Model \citep{worldmodels} & $906 \pm 21$ \\
    \end{tabular}
    \caption{\label{tab:final_compare}  CarRacing-v0 approaches with mean scores over 100 trials. Our approaches are shown in bold. }
\end{table}

\section{Discussion and Future Work}

While the agent reaches a decent score, it does fail occasionally. It usually happens when the agent is unable to correct itself due to loss of friction during turns at sharp corners with high speed. Compared to the expert model that enacts conservative driving-behavior, the current iterative model prefers more risky driving at high speed. Possibly, the expert policy has learned to slow down at corners, which helps maximize the reward. On the other hand, our planning agent does not seem to have explored sufficient rollouts of this kind to make the MD-RNN learn to associate higher rewards with slower driving when approaching corners.

Another issue occurs when the agent approaches the right corners. In many cases, the agent can complete right corners though there are times where the agent does not know whether to turn or not. In these scenarios, the agent usually brakes or slows down while trying to navigate the race track in a sensible direction. This phenomenon is likely due to the right turns being underrepresented in the generated tracks that are biased towards containing mainly left turns. Consequently, the MDN-RNN is unable to represent right turns in the simulated environment compared to other frequently occurring segments of the track. Arguably, both issues resolve by running more iterative training iterations. However, it also depends on how often the issues arise in the generated rollouts. Noteworthy, the issues occurred more often in the random model compared to the iterative model, which indicates that the iterative training procedure can be an effective method of improving the world model.
\section*{Acknowledgments}
We would like to thank Mathias Kristian Kyndlo Löwe for helping us with computational infrastructure. A special thanks go to Corentin Tallec and his team for providing the PyTorch open-source implementation of \textit{World Models} \cite{worldmodels}. We also thank Simon Lucas, Chris Bamford, and Alexander Dockhorn for helpful suggestions. This project was supported by a DFF Sapere Aude Starting Grant and by the Danish Ministry of Education and Science, Digital Pilot Hub and Skylab Digital.

\bibliographystyle{splncs04}

\bibliography{references}

\end{document}